\pdfoutput=1

\documentclass[letterpaper]{article} 
\usepackage[draft]{aaai2026}  
\usepackage{times}  
\usepackage{helvet}  
\usepackage{courier}  
\usepackage[hyphens]{url}  
\usepackage{graphicx} 
\usepackage{natbib}  
\usepackage{caption} 
\usepackage{algorithm}
\usepackage{algorithmic}

\usepackage{newfloat}
\usepackage{listings}
\DeclareCaptionStyle{ruled}{labelfont=normalfont,labelsep=colon,strut=off} 

\floatstyle{ruled}
\newfloat{listing}{tb}{lst}{}
\floatname{listing}{Listing}
\usepackage{booktabs}
\usepackage{multirow}
\usepackage{verbatimbox}
\usepackage{fancyvrb}
\usepackage[most]{tcolorbox}
\usepackage{subcaption} 

\title{ToolCritic: Detecting and Correcting Tool-Use Errors in Dialogue Systems}

\author {
    Hassan Hamad\textsuperscript{\rm 1}\thanks{work done during internship at Amazon.},
    Yingru Xu\textsuperscript{\rm 2},
    Liang Zhao\textsuperscript{\rm 2},
    Wenbo Yan\textsuperscript{\rm 2},
    Narendra Gyanchandani\textsuperscript{\rm 2}
}
\affiliations {
    \textsuperscript{\rm 1}Ming Hsieh Department of Electrical and Computer Engineering, University of Southern California, Los Angeles, California\\
    \textsuperscript{\rm 2}Amazon \\
    hhamad@usc.edu, yingruxu@amazon.com, confd@amazon.com, yanwenb@amazon.com, gyanchan@audible.com
}

\begin{document}

\maketitle

\begin{abstract}
Tool-augmented large language models (LLMs) are increasingly employed in real-world applications, but tool usage errors still hinder their reliability. We introduce ToolCritic, a diagnostic framework that evaluates and improves LLM behavior in multi-turn, tool-augmented dialogues. ToolCritic detects eight distinct error types specific to tool-calling (e.g., premature invocation, argument misalignment, and misinterpretation of tool outputs) and provides targeted feedback to the main LLM. The main LLM, assumed to have strong reasoning, task understanding and orchestration capabilities, then revises its response based on ToolCritic’s feedback. We systematically define these error categories and construct a synthetic dataset to train ToolCritic. Experimental results on the Schema-Guided Dialogue (SGD) dataset demonstrate that ToolCritic improves tool-calling accuracy by up to 13\% over baselines, including zero-shot prompting and self-correction techniques. This represents a promising step toward more robust LLM integration with external tools in real-world dialogue applications.
\end{abstract}

\begin{figure*}[t]
\centering
  \includegraphics[width=\linewidth]{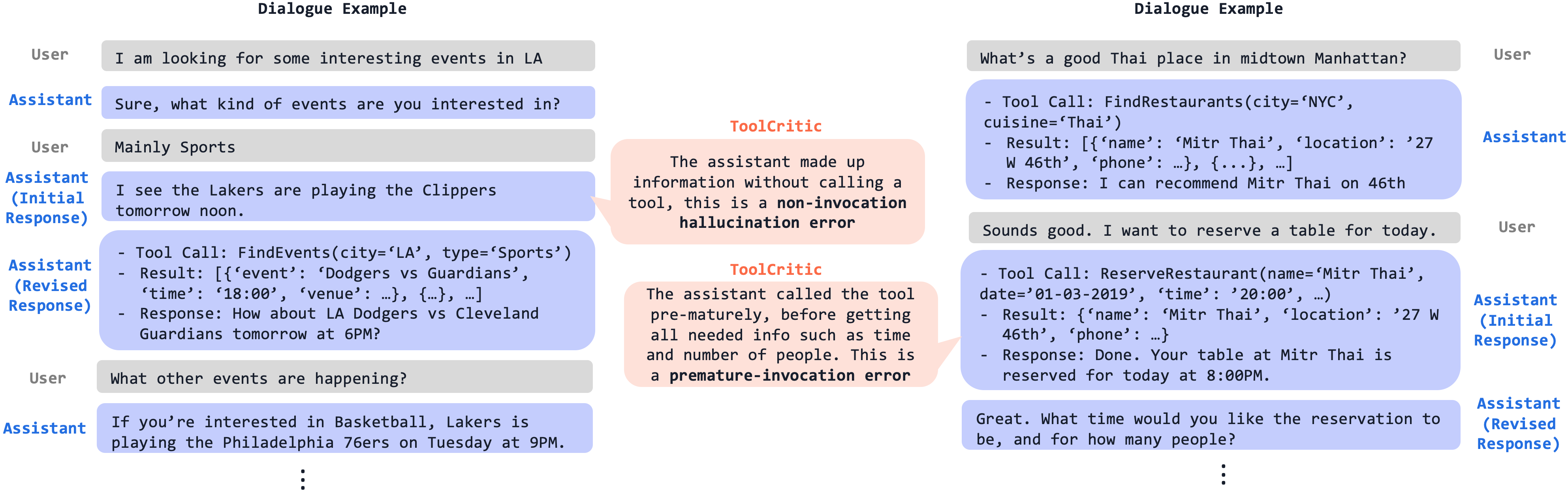}
  \caption {ToolCritic evaluates every assistant response in a conversation, whether a tool was called or not (this is not shown in the diagram due to space limitations). If ToolCritic detects an assistant error, it will produce a reasoning ``thought'' which will be fed back to the assistant. The assistant then will revise its own response based on the received feedback and produce an updated response. If ToolCritic does not detect a mistake, the conversation continues without interference. Note that while we show the Assistant ``Initial Response'' for clarity, in a real-world scenario, this response will be hidden from the user and only the revised response will be used.}
  \label{fig:examples}
\end{figure*}

\section{Introduction}

Large language models (LLMs) have made significant strides, becoming viable options for conversational agents. Beyond general-purpose conversations, there is a growing demand for these models to invoke external tools—such as booking APIs or search engines—to complete diverse task-oriented requests. This integration, often referred to as function-calling or tool-calling, significantly expands the scope of an LLM’s potential applications~\cite{DBLP:conf/acl/HuangZLZGLHZWSJ24, DBLP:journals/corr/abs-2405-17935}. Yet, reliable and accurate tool usage remains a major challenge: while modern LLMs can call external APIs, they frequently misidentify which tool to invoke, supply incorrect arguments, or even misinterpret the tool’s output~\cite{Qu2025-survey}. Function-calling has gained increasing attention in the development of foundational LLMs~\cite{llamma31, mistrallarge2, GPT4, claude3}. Still, as our results will show, even LLMs equipped with function-calling capabilities often struggle with effectively using tools in general.

LLMs are known to be prone to hallucinations and reasoning mistakes~\cite{DBLP:conf/ijcnlp/BangCLDSWLJYCDXF23, DBLP:journals/tacl/GuerreiroAWHBCM23, DBLP:journals/csur/JiLFYSXIBMF23}. Recent strategies to improve LLM reliability often center on intrinsic self-correction (where the LLM inspects and corrects its own reasoning)~\cite{DBLP:conf/nips/MadaanTGHGW0DPY23, DBLP:conf/nips/ShinnCGNY23}, or external feedback loops (where an external model or source identifies mistakes or provides feedback to the LLM)~\cite{criticgpt, CRITIC}. While self-correction shows promise, it remains inconsistent—LLMs often fail to identify their own errors reliably, leading to only incremental improvements~\cite{DBLP:conf/iclr/0009CMZYSZ24, tyen-etal-2024-llms}.

While these previous studies have focused on general reasoning errors, our work focuses on the function-calling ability of LLMs in multi-turn task-oriented dialogue (TOD) applications. Utilizing LLMs in multi-turn TOD scenarios, such as MultiWOZ~\cite{budzianowski2018large}, SGD~\cite{sgd}, and ToolTalk~\cite{tooltalk} datasets, presents unique challenges. At each turn, the LLM must determine whether a tool call is necessary, which requires tracking user intent over multiple turns, identifying the correct moments for tool invocation, selecting the appropriate tool from a potentially large set, and accurately using the tool’s specifications. This setup is prone to a wide range of errors and failure modes, more so than in general conversational dialogue. In this work, we identify and categorize eight distinct error types related to tool usage, such as premature tool invocation—where an LLM calls a tool before gathering all necessary information from the user—and observation-reasoning errors, where the LLM correctly calls a tool and receives an output but misinterprets this output, leading to an incorrect response.

Motivated by findings that LLMs struggle with self-correction, we propose \textbf{\emph{ToolCritic}}: a model that detects errors in LLM tool usage during multi-turn TODs and provides descriptive feedback that the LLM can use to revise and improve its responses (see examples in Figure~\ref{fig:examples}). Our approach works as follows: In a dialogue between a user (human) and an assistant (an LLM), at each turn, ToolCritic inspects the assistant’s proposed response. If an error is detected, ToolCritic provides targeted, textual feedback describing the mistake. The assistant LLM, assumed to have strong in-context learning capabilities, uses this feedback to revise its response and correct the error on-the-fly. Our results confirm that self-correction is not viable in this context and may even degrade performance.

Unlike previous work that broadly examines LLM reasoning errors or focuses on the function-calling ability of LLMs in isolation, we specifically investigate LLM function-calling in conversational dialogue. Our contributions can be summarized as follows:

\begin{itemize}
    \item An \textbf{error taxonomy} of eight distinct failure modes for LLMs calling external tools in multi-turn dialogues.
    
    \item A \textbf{diagnostic model (ToolCritic)} trained on a synthetic dataset to detect errors and provide descriptive feedback for the LLM-as-Tool architecture in conversational dialogue.

    \item A feedback loop mechanism that demonstrates up to a $13\%$ improvement in dialogue success rates over self-correction and zero-shot approaches on the Schema Guided Dialogue (SGD) dataset~\cite{sgd}.
    
\end{itemize}

The remainder of this paper is structured as follows: Section~\ref{sec:background} provides some background and reviews related work to highlight the differences from our approach. Section~\ref{sec:dataset} describes the pipeline to generate the synthetic dataset needed for fine-tuning. Section~\ref{sec:model} discusses the training of ToolCritic. Section~\ref{sec:experiments} presents the experimental results. We conclude in Section~\ref{sec:conclusion}.

\begin{figure*}[t]
\centering
  \includegraphics[width=0.75\linewidth]{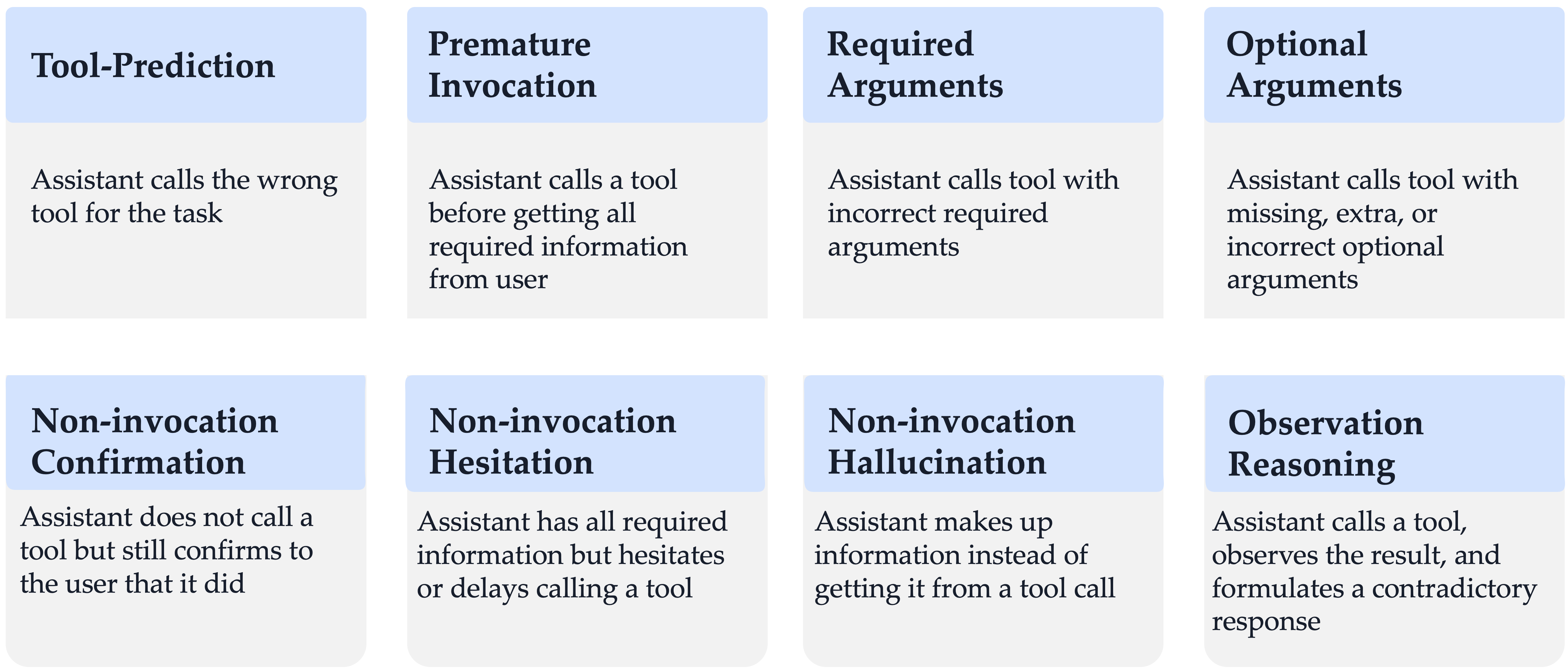}
  \caption {Each error category defines a specific mistake, or failure mode, that is common for LLMs when interacting with tools in multi-turn dialogue. This granularity ensures the resulting diagnostic model, ToolCritic, can provide very specific feedback for the assistant LLM.}
  \label{fig:errors}
\end{figure*}

\section{Background and Related Work}\label{sec:background}

The integration of tools with large language models (LLMs), also referred to as function-calling, has gained significant attention in recent years, driven by the enhanced reasoning capabilities of these models. Since LLMs are primarily pretrained on language modeling tasks, additional post-training is helpful to improve their function-calling abilities. Toolformer~\cite{toolformer} employs a self-supervised training approach to enhance a model's proficiency in tool invocation. In~\cite{gorilla}, the authors developed a large-scale dataset of user prompts and API calls, followed by self-instruct fine-tuning of a LLaMA-7B-based model, which significantly improved API invocation accuracy.

As discussed in the introduction, LLMs are prone to reasoning errors and hallucinations~\cite{DBLP:conf/ijcnlp/BangCLDSWLJYCDXF23, DBLP:journals/tacl/GuerreiroAWHBCM23, DBLP:journals/csur/JiLFYSXIBMF23}, and these issues are often exacerbated in tool-calling scenarios~\cite{Qu2025-survey}. While self-correction techniques~\cite{DBLP:conf/nips/MadaanTGHGW0DPY23, DBLP:conf/nips/ShinnCGNY23}—where an LLM is provided with its own response for improvement—have shown potential, they have also faced criticism that raises doubts about their effectiveness. For example,~\cite{DBLP:conf/iclr/0009CMZYSZ24} attributed much of the improvement in earlier self-correction work to poor baselines, unfair comparisons, and suboptimal prompt design. Another study~\cite{tyen-etal-2024-llms} further confirmed these concerns, demonstrating that LLMs struggle with self-correction because they are not adept at identifying their own errors—an essential first step in self-correction. The same study also revealed that when errors are explicitly pointed out, LLMs are more capable of self-correcting.

In our experiments, we observe that LLMs struggle to identify their own mistakes when interacting with tools. Our findings support this prior research which suggests that external feedback is needed for correcting such errors. This conclusion has also been supported by new approaches. For instance, in~\cite{criticgpt}, the authors address the challenge of detecting bugs in LLM-generated code by training a GPT-4 model via reinforcement learning with human feedback (RLHF) on a dataset of user questions and buggy code. This model demonstrated significant improvements in bug detection compared to having the base GPT-4 model detect its own bugs. Another approach, the CRITIC framework~\cite{CRITIC}, involves LLMs using external tools to receive feedback for self-correction, such as performing a Google search to do fact-checking. 

Unlike these studies, our what distinguishes our work is that we focus more specifically on enhancing the fundamental tool-calling ability of LLMs within the context of conversational dialogue by utilizing a single diagnostic model. Multi-turn dialogues present a more complex challenge, requiring the LLM to invoke different tools at different turns, track user intent throughout the dialogue, interact with users to gather additional information or confirmations, and manage a dynamic conversational flow.

A closely related work is T-Eval~\cite{chen-etal-2024-eval}, which evaluates the tool utilization capabilities of LLMs by dividing the task into multiple sub-processes and assessing different models using a constructed dataset. However, T-Eval is purely an evaluation study, whereas our work aims to develop a model that can automatically evaluate and correct tool usage errors in real-time.

To investigate this problem, it is essential to examine datasets of conversational dialogues that involve tool-calling, particularly those focused on task-oriented dialogues. Examples include MultiWOZ~\cite{budzianowski2018large}, SGD~\cite{sgd}, and TaskMaster~\cite{byrne-etal-2019-taskmaster}. While these datasets can simulate tool-calls through mock interfaces, others, such as API-Bank~\cite{api-bank}, ToolTalk~\cite{tooltalk}, and Lucid~\cite{lucid}, implement actual tool-calls within the dialogues, for example, via a python backed. For our study, we focus on the SGD dataset due to its large scale aspect and the diverse range of tools it includes.

\section{Dataset Definition}\label{sec:dataset}

\subsection{Error Categories Definition}
To build a robust diagnostic system, we began by categorizing the common mistakes LLMs make during tool usage in conversations. Our literature review identified failures in instruction following, reasoning, and planning as the root causes. We divided these into eight specific error types (see Figure~\ref{fig:errors}), enabling fine-grained error detection and actionable feedback. Detailed definitions and examples for each category are included in Appendix~\ref{app:error-categories}. Two key points need to be highlighted:

\textbf{Importance of Error Granularity.} Consider for example the ``required arguments'' and ``optional arguments'' categories. Errors in required arguments usually stem from failing to capture user intent or aligning input with tool specifications. Optional arguments, on the other hand, can indicate mismanagement of non-critical parameters. Although these could be grouped together, distinguishing between these error types helps the model understand the nuances of each failure mode, resulting in more effective error detection and correction.
    
\textbf{Interrelated Errors.} Multiple errors often co-occur within a single turn. For example, a ``premature tool'' invocation might cascade into missing required or optional arguments. ToolCritic is designed to identify and prioritize the primary error only, ensuring feedback is efficient and avoids redundant corrections.

\subsection{Synthetically-Modified Error Dataset}
\paragraph{Dialogue Notation.}
We define a multi-turn dialogue \(d\) as a sequence of user--assistant turns:
\[
d = \bigl[(u_1, a_1), (u_2, a_2), \dots, (u_N, a_N)\bigr],
\]
where \(u_i\) is the user’s \(i\)-th utterance, \(a_i\) is the assistant’s \(i\)-th response, and \(N\) is the total number of turns. Given that the effectiveness of ToolCritic is dependent on the dataset, we selected the \textbf{Schema-Guided Dialogue (SGD) dataset}~\cite{sgd} for this study because of its large size and diversity in tool calls. Appendices~\ref{app:error-categories} and~\ref{app:generation-prompt} include some examples from this dataset. In total, $40$ different tools from the dataset are used in our experiments. To train ToolCritic effectively, we require both negative (error-free dialogues) and positive (error-injected dialogue) examples:

\textbf{Error-Free Dialogues.}
An \emph{error-free} dialogue is one in which the assistant’s behavior—especially its tool calls—is fully correct and aligned with the user’s intent. We obtain examples directly from the SGD dataset, as it already consists of dialogues with correct tool usage.

\textbf{Error-Injected Dialogues.} An \emph{error-injected} dialogue is one containing ``improper'' tool usage. We construct these \emph{positive} examples by injecting a single mistake into one of the assistant turns of an error-free dialogue. Formally, for a chosen turn index \(k\), we modify \(a_k\) to introduce a specific error type (e.g., calling a tool prematurely, missing a required argument). This yields a new dialogue
\[
d^\prime = \bigl[(u_1,a_1),\dots,(u_{k},a_{k}^{\prime})\bigr],
\]
where \(a_{k}^{\prime}\) is an erroneous response and we discard the remaining turns.

\textbf{Few-Shot Prompting.} To generate the \emph{error-injected} dialogues, we first manually write some few seed examples (five to seven per error category) that illustrate the eight errors we aim to detect (e.g., “premature invocation,” “missing required arguments”). These examples were carefully designed to span various domains and encompass different failure scenarios within each error category. We then apply few-shot prompting with the \textbf{Claude 3.5 Sonnet} model, by providing the seed examples and one new error-free dialogue from the SGD dataset while instructing it to insert one of the eight errors at a randomly chosen turn in the dialogue and produce the appropriate labels. These modified dialogues simulate real-world mistakes that an assistant might make when interacting with tools. Each modified dialogue was annotated with: the error category, indicating the type of mistake, and a reasoning thought, explaining why the response was incorrect and how the error can be identified. In total, we generated $300$ examples per error category, resulting in $2,400$ error-injected dialogues. Details explaining this data generation process and the prompt templates used are provided in Appendix~\ref{app:generation-prompt}.

\textbf{Data Quality}. During data creation, we adopted an iterative quality-control process. For each error category, we randomly sampled $10$ generated dialogues for manual inspection. If more than one dialogue was found not to follow the error definition, we refined the prompt and regenerated the data. This manual check ensured a high quality synthetic data with no more than $10\%$ noisy labels.

\begin{figure*}[ht]
    \centering
    \begin{subfigure}[b]{0.44\textwidth} 
        \centering
        \includegraphics[width=\textwidth]{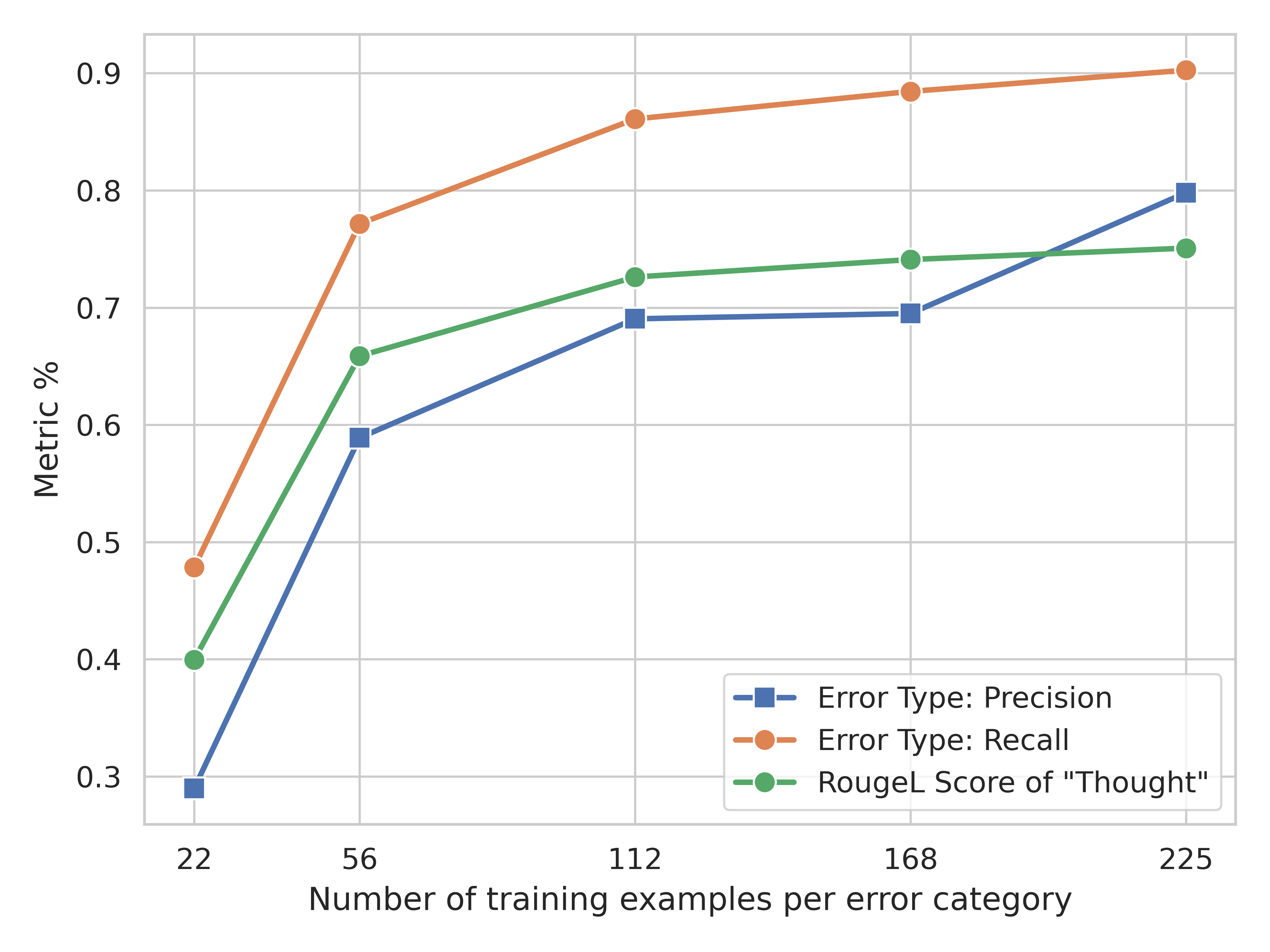} 
        \caption{Precision, Recall and ROUGE score results when training ToolCritic on different sizes of the training data.}
        \label{fig:sub-a}
    \end{subfigure}
    \hspace{0.1\textwidth} 
    \begin{subfigure}[b]{0.44\textwidth}
        \centering
        \includegraphics[width=\textwidth]{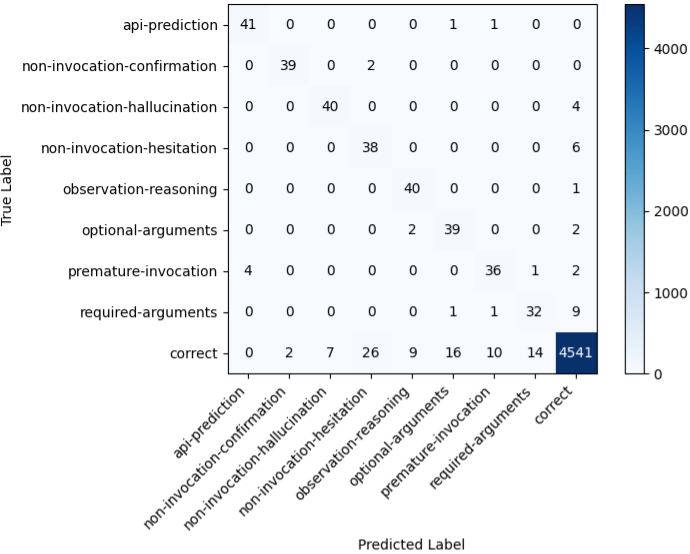} 
        \caption{Confusion matrix of ToolCritic on $D^{\text{test}}$ when using all training data, i.e. 225 examples per error category}
        \label{fig:sub-b}
    \end{subfigure}
    \caption{Evaluation results of ToolCritic on the test data $D^{\text{test}}$.}
    \label{fig:curves}
\end{figure*}

\section{ToolCritic}\label{sec:model}
In this section, we describe how ToolCritic was trained and we report the evaluation metrics on the generated synthetic dataset.

\subsection{Fine-Tuning \& Evaluation Setup}\label{sec:finetune}

\textbf{Data Splits.} We generated a total of $2,\!400$ error-injected dialogues and collected an equal number of error-free dialogues, resulting in a balanced dataset of $4,\!800$ dialogues from the SGD dataset. This dataset is divided into training $D^{\text{train}}$, evaluation $D^{\text{eval}}$, and test sets $D^{\text{test}}$ with a $70\%$, $15\%$, and $15\%$ stratified split, respectively. This results in a training dataset comprising $225$ examples for each error category and an equal number of error-free examples, totaling $3,600$ dialogues.

Each data point in $D^{\text{train}}$ is represented as a pair $(d, y)$, where:
\begin{itemize}
    \item $d$ is either a randomly subsampled subdialogue from an error-free dialogue or the complete dialogue from an error-injected dialogue.
    \item $y$ is the label, which is \texttt{"no error"} for error-free dialogues or \texttt{"<error category>: <thought>"} for error-injected dialogues.
\end{itemize}

This construction ensures two main points: First, the training set is balanced, with an equal number of positive (error-injected) and negative (error-free) examples. Second, the model is trained on complete and incomplete dialogues, of varying lengths. This ensures that, during test time, ToolCritic doesn't expect a complete dialogue as input but can be invoked at any part in the conversation. 

For evaluation and testing, we adopt a "roll-out" approach to simulate real-world scenarios where errors are sparse. Specifically, each dialogue with $K$ turns is transformed into $K$ data points. Each data point $(d_k, y_k)$ consists of:
\begin{itemize}
    \item $d_k$: The subdialogue containing all turns up to and including turn $k$.
    \item $y_k$: The label, which for error-injected dialogues is \texttt{"no error"} for the first $K-1$ datapoints $d_1, d_2, .., d_{K-1}$ and is \texttt{"<thought>: <error category>"} for the last datapoint $d_K$. For error-free dialogues, the label is \texttt{"no error"} for all $K$ datapoints.
\end{itemize}

This method ensures that during evaluation and testing, the majority of data points are negative (\texttt{"no error"}), reflecting the low occurrence of errors in real-world applications. Consequently, the evaluation metrics can more accurately assess the model's ability to detect rare errors amidst predominantly correct responses.

\textbf{Taining Details.} We fine-tuned an open-source language model, specifically \textbf{LLaMA 3.1 8B instruct}, on $D^{\text{train}}$ for $5$ epochs with a learning rate of 1e-5 and a batch size of $2$. Full fine-tuning was performed, meaning all model weights were updated using bfloat16 precision on a p4.2xlarge instance with 8 Nvidia A100 GPUs. Model weights were updated based only on the reasoning output, not the dialogue itself~\footnote{See the "DataCollatorForCompletionOnlyLM" from the trl package \url{https://huggingface.co/docs/trl/en/sft_trainer}}. Additional training details and the prompts used for fine-tuning are provided in Appendix~\ref{app:training}.

\textbf{Evaluation Metrics.} We adopt precision and recall to evaluate \emph{error detection}, while the ROUGE score measures the quality of the generated \emph{reasoning text} when an error is flagged.

\textbf{Training-Data Size Variation.} Additionally, we train our model on $\{10\%, 25\%, 50\%, 75\%, 100\%\}$ of the training data $D^{\text{train}}$ to investigate whether additional labeled dialogues can yield further improvements.

\subsection{ToolCritic Evaluation Results on Synthetic Dataset}
Figure~\ref{fig:curves} presents the evaluation results. The results show a clear improvement in all metrics with the addition of more training data. The model achieves a high recall of approximately $90\%$, though precision lags behind at around $80\%$. This discrepancy is expected given that the test dataset contains significantly more negative data (error-free dialogues) than positive data (dialogues with errors). In real-world applications, precision may be more critical to minimize false positives, and the appropriate threshold will depend on the specific use case. 

As training data increases, recall and ROUGE scores approach saturation, while precision shows more room for improvement, suggesting that larger datasets may still help reduce false positives in real-world applications.

\noindent\textbf{Error Type Analysis}
The confusion matrix, shown in Figure~\ref{fig:sub-b}, provides insight into the model's performance across different error categories when trained on the entire training dataset. The results indicate similar performance across most error categories, with some minor variations. We also observe the distribution of false positives, with the ``non-invocation hesitation'' category being the most common. Further analysis is required to understand how to mitigate this effect. In Appendix~\ref{app:confsuion}, we also present the confusion matrix for the low-label regime, where the model was trained with only $10\%$ of the training data. This analysis helps identify which error categories are more challenging when training data is limited, providing a clearer view of how dataset size affects performance across specific error types.

\section{Experiments on Real LLM Mistakes}\label{sec:experiments}

In the previous section, we introduced ToolCritic, a diagnostic model that displayed strong performance in detecting synthetically generated LLM tool-calling errors. This section focuses on testing ToolCritic's ability to detect and correct real-world, not synthetic, LLM errors by evaluating multiple LLMs on the unseen test split of the Schema-Guided Dialogue (SGD) dataset. 

\begin{figure}[h!]
  \includegraphics[width=0.95\linewidth]{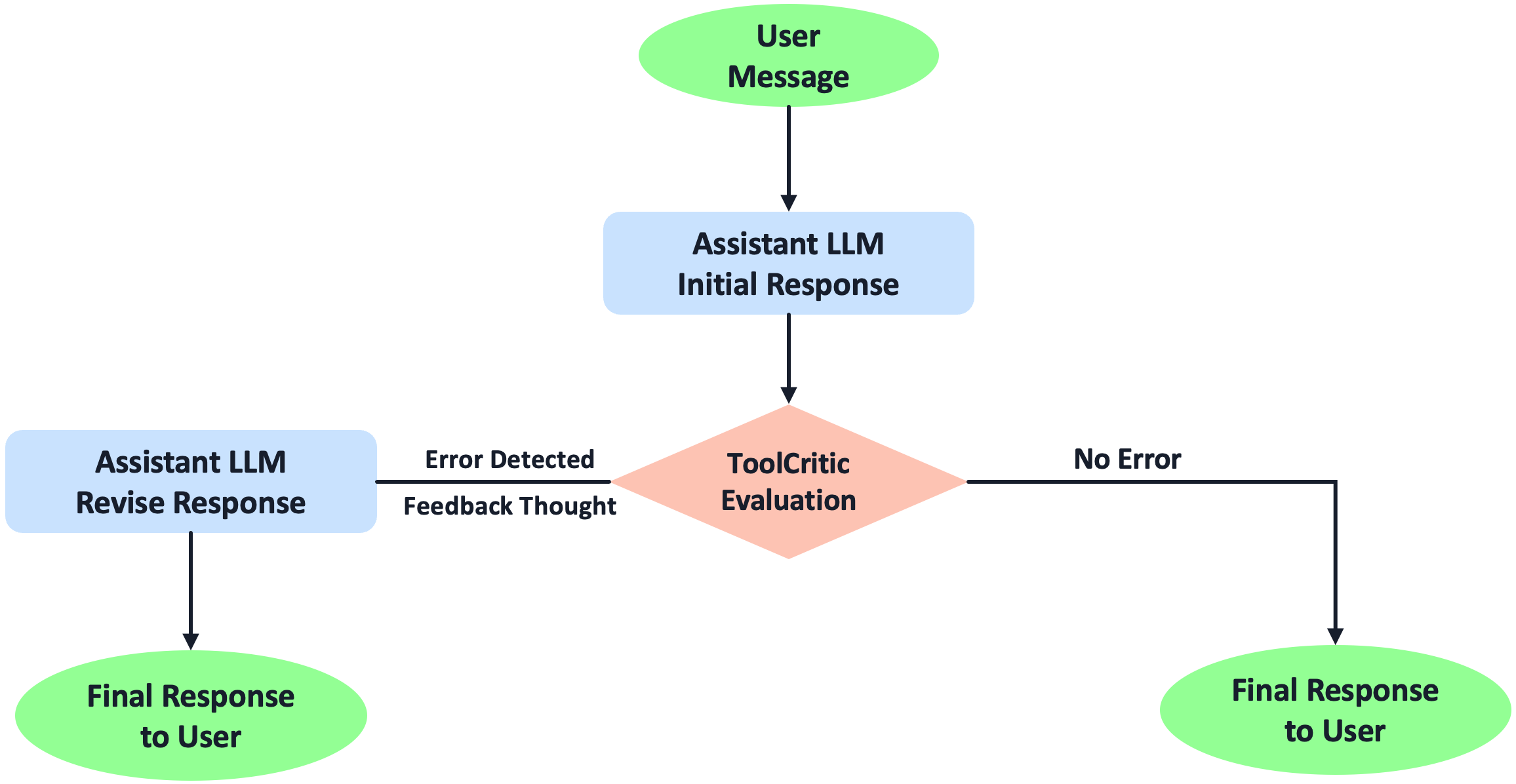}
  \caption {Flowchart illustrating the evaluation process for each turn in the dialogue. ToolCritic evaluates the LLM's response, providing feedback for correction if an error is detected. Note that the LLM performs a single attempt at correction and the revised response is passed to the user without additional evaluation. This allows us to balance accuracy with inference and latency cost.}
  \label{fig:flowchart}
\end{figure}

\begin{table*}[htbp!]
\footnotesize
    \centering
    \begin{tabular}{llrrrrr}
        \toprule
        {\textbf{Assistant LLM}} &
        \multicolumn{1}{c}{\textbf{Evaluation}} &
        {\textbf{Success Rate}} &
        {\textbf{Incorrect}} &
        {\textbf{Precision}} &
        {\textbf{Recall} } \\
        {} &
        \multicolumn{1}{c}{\textbf{Scenario}} &
        {\textbf{(\%)} \textuparrow} &
        {\textbf{Action Rate (\%) \textdownarrow}} &
        {(\%) \textuparrow} &
        {(\%) \textuparrow} \\
        \midrule
        \multirow{4}{*}{LLaMA 3.1 70B} & Baseline (zero-shot) & $5.46^{\pm 5.46}$  & $84.47^{\pm 7.83}$ & $21.51^{\pm 4.11}$ & $40.68^{\pm 7.34}$ \\
        
        & Self-Correction & $3.90^{\pm 3.90}$  & $84.93^{\pm 7.36}$ & $24.06^{\pm 2.32}$ & \bfseries 45.24$^{\pm 3.57}$ \\
        
        & ToolCritic (error-only feedback) & $3.90^{\pm 3.90}$  & $84.13^{\pm 8.17}$ & $24.91^{\pm 3.86}$ & $40.68^{\pm 7.34}$ \\
        
        & ToolCritic (full feedback)$^{1}$ & \bfseries 10.15$^{\pm 10.15}$  & \bfseries 77.32$^{\pm 11.56}$ & \bfseries 33.03$^{\pm 4.46}$ &  $43.83^{\pm 10.49}$ \\
        
        \midrule
        \multirow{4}{*}{Mistral Large 2} & Baseline (zero-shot) & $7.35^{\pm 0.17}$  & $91.85^{\pm 0.11}$ & $18.54^{\pm 0.18}$ & $34.26^{\pm 0.45}$ \\
        
        & Self-Correction & $6.26^{\pm 0.18}$  & $92.02^{\pm 0.03}$ & $17.05^{\pm 0.19}$ & $35.58^{\pm 0.17}$ \\
        
        & ToolCritic (error-only feedback) & $8.16^{\pm 0.97}$  & $91.64^{\pm 0.15}$ & $19.62^{\pm 0.18}$ & $36.09^{\pm 0.22}$ \\
        
        & ToolCritic (full feedback) & \bfseries 10.61$^{\pm 0.94}$  & \bfseries 84.74$^{\pm 0.61}$ & \bfseries 26.90$^{\pm 0.45}$ & \bfseries 50.24$^{\pm 0.55}$ \\
        
        \midrule
        \multirow{4}{*}{Claude 3 Sonnet} & Baseline (zero-shot) & $14.15^{\pm 0.30}$  & $55.68^{\pm 0.02}$ & $44.28^{\pm 0.24}$ & $64.60^{\pm 0.21}$ \\
        
        & Self-Correction & $16.62^{\pm 0.58}$  & $54.04^{\pm 0.03}$ & $46.31^{\pm 0.02}$ & $66.62^{\pm 0.59}$ \\
        
        & ToolCritic (error-only feedback) & $19.95^{\pm 0.35}$  & $52.63^{\pm 0.13}$ & $50.21^{\pm 0.10}$ & $67.66^{\pm 0.06}$ \\
        
        & ToolCritic (full feedback) & \bfseries 27.88 $^{\pm 0.85}$  & \bfseries 46.27$^{\pm 0.03}$ & \bfseries 57.60$^{\pm 0.08}$ & \bfseries 76.78$^{\pm 0.23}$ \\
        \bottomrule
    \end{tabular}
    
    \caption{\label{tab:results} Main results of the experimental evaluation on the SGD test data. Each experiment setup is repeated twice with random seeds. The mean and standard deviation (in superscript) are reported for each metric.}
\end{table*}

\subsection{Evaluation Setup}

Evaluating LLMs in multi-turn dialogues, particularly in the context of tool usage, presents significant challenges. Traditional evaluation methods, such as tracking dialogue state or metrics like inform rate, are becoming less effective given the improved capabilities of modern LLMs. Ideally, a robust evaluation would involve live interactions between the LLM and human users, followed by an assessment of the LLM’s responses. However, such an approach is impractical due to the high cost and complexity of involving human evaluators.\\
A more feasible alternative is to use static user utterances from a dataset and have the LLM generate responses as the assistant. However, this method risks producing incoherent conversations, as the assistant’s responses may diverge from the static dataset, creating a mismatch between the context and the assistant’s output. To address this issue,~\citet{xu-etal-2024-rethinking} proposed an evaluation setup where an LLM acts as both the assistant and the user, i.e. a ``user simulator'', with another LLM serving as a judge to evaluate the responses. While this setup allows for a dynamic evaluation, it assumes that the LLM functioning as the judge can reliably detect errors, which contradicts our premise that LLMs often struggle to identify their own mistakes.\\
To mitigate these challenges, we adopt the evaluation setup introduced in~\cite{tooltalk}, which is tailored for assessing tool-augmented LLMs in dialogues. In this setup, the LLM is provided with the ground truth conversation history at each turn, rather than its own previous responses, and is tasked with predicting the next response. This approach (sometimes referred to as ``teacher forcing'') helps avoid incoherence issues, making it more suitable for evaluating LLMs' tool-calling abilities. At each turn, the LLM’s predicted response is fed into ToolCritic, which checks for any mistakes. If ToolCritic detects an error, it provides reasoning feedback, and the LLM revises its response accordingly (see Figure~\ref{fig:flowchart}).

\subsection{Models and Benchmarks}
\textbf{LLMs as assistant.} We evaluated three different LLMs as the assistant: Claude 3 Sonnet~\cite{claude3}, a closed-source model, and two open-source models, Mistral Large 2~\cite{mistrallarge2} and LLaMA 3.1 70B Instruct~\cite{llamma31}. All three models have tool-calling capabilities. The temperature is fixed to $0.1$ for both ToolCritic and the LLMs.

\textbf{Experiment Setup}. For each assistant LLM, we compared four different experimental setups:

\begin{itemize}
    \item Baseline (zero-shot): The LLM generates a response without any correction or feedback.
    \item Self-correction: The LLM reviews and attempts to correct its own responses following the Reflexion method~\citet{DBLP:conf/nips/ShinnCGNY23}.
    \item ToolCritic full feedback: The LLM receives external feedback from ToolCritic on detected errors, and revises its responses accordingly.
    \item ToolCritic error-only feedback: An ablation study where the LLM only receives the general description of the predicted error from ToolCritic, from Figure~\ref{fig:errors}, but not the ``reasoning thought'', and revises its responses accordingly. For example, feedback might indicate a tool argument error but doesn't contain a reasoning to indicate which argument. This allows us to assess the importance of ToolCritic's detailed feedback.
\end{itemize}

\textbf{Metrics.} We adopted the metrics defined in~\cite{tooltalk}, which include precision, recall, and incorrect action rate. An ``action tool'' refers to any tool that triggers real-world changes, such as setting an alarm or booking a reservation, as opposed to passive tools like information retrieval. A tool call is considered correct if all arguments match the ground truth, with fuzzy matching used for free-form string arguments. Additionally, we monitored a dialogue-level metric called ``success rate'', which measures the assistant’s ability to correctly predict all necessary tool calls without making unnecessary action tool calls (extra passive tool calls are allowed). Each experiment was repeated twice with random seeds, and we report the mean and standard deviation for each metric.

\subsection{Results}
The experimental results are summarized in Table~\ref{tab:results}.

\noindent\textbf{Baseline.}  Baseline performance varies across the LLMs, with Claude 3 Sonnet achieving the highest scores of $14.15\%$, $44.28\%$ and $64.60\%$ on success, precision, and recall, respectively. These numbers highlight the difficulty of tool usage in multi-turn dialogues, even for advanced models. LLaMA 3.1 had the lowest baseline performance, with a success rate of just $5.46\%$.

\noindent\textbf{Self-Correction.} Results of the self-correction experiments reinforce that LLMs struggle to detect their own mistakes. In many cases, self-correction produced only minor improvements, and in some instances, even degraded performance.

\noindent\textbf{ToolCritic Feedback.} Using full feedback from ToolCritic resulted in significant improvements across all models. Claude 3 Sonnet saw the greatest improvement, with a success rate increase of $13\%$. The gains were smaller for Mistral and LLaMA, likely due to their relatively poor performance in tool usage, suggesting that these models still struggle to effectively use tools in a dialogue context.

\noindent\textbf{Ablation.} We observed smaller improvements when the feedback consists of generic error descriptions only, i.e. no reasoning. This suggests that while simple error descriptions can be helpful, more complex mistakes require nuanced feedback for the LLM to effectively correct itself. For example, if the assistant calls a tool prematurely, an error description indicating a ``premature invocation'' may suffice. However, if the error involves a tool argument, detailed reasoning is necessary to guide the LLM toward the specific argument.

\subsection{Error Analysis}
\begin{figure}[t]
  \includegraphics[width=0.9\linewidth]{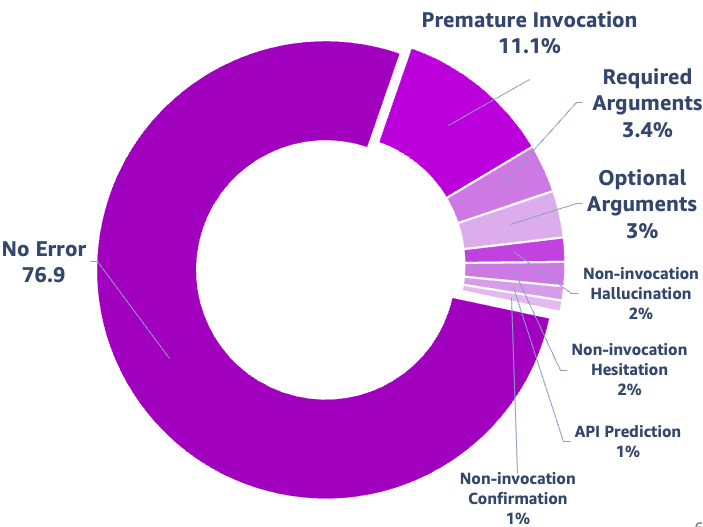}
  \caption {``Error Profile'' of Claude 3 Sonnet. The diagram shows the distribution of mistakes detected by ToolCritic when Claude 3 Sonnet serves as the assistant on the SGD test set.}
  \label{fig:donut}
\end{figure}

\textbf{Error Profile.} Figure~\ref{fig:donut} shows the distribution of ToolCritic-detected errors when evaluating Claude 3 Sonnet on the SGD test set. In 77\% of turns, no errors were detected. ``Premature invocation'' was the most common error, aligning with findings in~\cite{tooltalk}. Tool argument errors were the next most frequent error. Such error profiles can help identify model weaknesses and guide improvements.

\noindent\textbf{Human Evaluation.} Beyond tool correctness, we evaluated ToolCritic’s feedback quality via a turn-level human study on $100$ dialogues with Claude 3 Sonnet. Each turn was labeled into one of six outcomes: (1) useless/no difference, (2) made correct, (3) made better, (4) made incorrect, (5) missed an error, or (6) caught error but couldn't correct. ToolCritic with full feedback corrected 26.67\% of responses and improved another 26.67\%, outperforming self-reflection and error-only feedback. The study also validated our error taxonomy's completeness, as no uncategorized errors were found. Full details are in Appendix~\ref{app:human-study}.

\subsection{Generalization}\label{sec:generalization}
Results on the SGD dataset show a clear improvement in tool-calling accuracy with ToolCritic's feedback. A natural question is whether ToolCritic generalizes well. We distinguish between two forms of generalization:

\textbf{Generalization to new error types:} Our model is trained to detect and reason over eight well-defined categories of tool usage errors. While no taxonomy can be truly exhaustive, we found our error categories to be highly comprehensive in practice. In our human evaluation study (Appendix~\ref{app:human-study}), all errors made by the assistant LLMs were successfully captured by the defined categories, and no additional error types were identified—suggesting strong coverage of real-world tool-related mistakes.

\textbf{Generalization to new tools:} ToolCritic is trained on the tools and schemas from the SGD dataset and is designed to operate within that tool set, without assuming zero-shot generalization to unseen APIs. Nevertheless, to explore the model’s flexibility, we include illustrative examples in Appendix~\ref{app:ood} where ToolCritic—without retraining—is applied to examples from the ToolTalk dataset~\cite{tooltalk}. We emphasize that \textbf{ToolCritic is primarily designed to operate within the same tool set it was trained on, and we do not claim generalization to unseen tool definitions or APIs}. Nevertheless, these examples demonstrate that ToolCritic exhibits some capacity to identify high-level error patterns even for tools outside its training distribution. Future work could strengthen this capability via approaches like tool description abstraction, or reinforcement learning across various tool schemas.

\section{Conclusion}\label{sec:conclusion}

We introduced ToolCritic, a diagnostic model designed to enhance tool usage in multi-turn dialogues by providing actionable feedback on LLM responses. By identifying errors and providing real-time feedback, our approach improved tool-calling accuracy and dialogue success rates, especially when using the full feedback from the diagnostic model. Though effective, this method introduces challenges such as higher inference costs and the need for labeled training data. Future work will focus on addressing these limitations and expanding the model's capabilities to further improve tool-augmented LLMs in task-oriented dialogue applications.

\section{Limitations}\label{sec:limitations}

\textbf{Inference Cost}. Using ToolCritic adds latency: one critic inference per turn, plus an additional assistant LLM call if an error is found. This can be costly, especially in real-time use. Future work could explore reducing frequency of calls or compressing the assistant further.

\textbf{Data Labeling Cost.} ToolCritic requires labeled dialogues for training. Although we generate errors synthetically, the pipeline still depends on high-quality, error-free dialogues. Future work may focus on fully synthetic or self-supervised alternatives.

\textbf{Generalization.} As mentioned in Section~\ref{sec:generalization}, ToolCritic is trained on a fixed set of tool schemas from the SGD dataset and is not designed to generalize zero-shot to entirely new tools. Supporting broader tool generalization remains an open direction for future work.

\textbf{Chained Tool Calls.} Our setup assumes a single tool call per turn, as in SGD. Real-world use cases may involve multiple tools per turn, which would require new error types and more complex reasoning to support.

\appendix


\bibliography{aaai2026}

\setcounter{secnumdepth}{2}

\section{Illustrative Examples of Error Categories}\label{app:error-categories}

This appendix provides examples to explain each error category introduced in the main paper. We start with a correct dialogue without any errors and then systematically modify the dialogue to simulate each error type, accompanied by a short explanation.

\subsection{Good Dialogue Without Errors}

Below is an example of a well-executed dialogue where the assistant performs correctly without any errors:

\begin{tcolorbox}[breakable, enhanced, colframe=black!75!white, colback=white, sharp corners]
\begin{lstlisting}[language={}]
# Turn 1
    USER: "I need to find a seat on a bus."
    ASSISTANT: "Where are you leaving from? Where are you going?"

# Turn 2
    USER: "I am leaving from Vancouver to go to Seattle."
    ASSISTANT: "When are you leaving?"

# Turn 3
    USER: "I am leaving on the 12th of this month. I need 1 ticket."
    ASSISTANT:
        - API CALL: FindBus(from_location='Vancouver', leaving_date='2019-03-12', to_location='Seattle', travelers='1')
        - RESULT:
            * {"fare": "29", "leaving_time": "06:40"}
            * {"fare": "31", "leaving_time": "08:10"}
        - RESPONSE: "I found multiple options. The first bus departs at 6:40 am for $29."

# Turn 4
    USER: "What is the station I am leaving from?"
    ASSISTANT: "The bus departs from Pacific Central Station."

# Turn 5
    USER: "Book the 6:40 am option."
    ASSISTANT:
        - API CALL: BookBus(from_location='Vancouver', leaving_date='2019-03-12', leaving_time='06:40', to_location='Seattle', travelers='1')
        - RESULT: {"status": "confirmed", "booking_id": "12345"}
        - RESPONSE: "Your ticket has been booked. Your booking ID is 12345."
\end{lstlisting}
\end{tcolorbox}

\subsection{Error Categories}

\subsubsection{Premature Invocation}
\textbf{Definition:} The assistant calls a tool before gathering all necessary information from the user.

\begin{tcolorbox}[breakable, enhanced, colframe=black!75!white, colback=white, sharp corners]
\begin{lstlisting}[language={}]
# Turn 1
    USER: "I need to find a seat on a bus."
    ASSISTANT: "Where are you leaving from? Where are you going?"

# Turn 2
    USER: "I am leaving from Vancouver to go to Seattle."
    ASSISTANT:
        - API CALL: FindBus(from_location='Vancouver', leaving_date='2019-03-15', to_location='Seattle', travelers='1')
        - RESULT:
            * {"fare": "32", "leaving_time": "07:30"}
            * {"fare": "45", "leaving_time": "08:50"}
        - RESPONSE: "I found multiple options. The first bus departs at 7:30 am for $32."
\end{lstlisting}
\end{tcolorbox}

\textbf{Explanation:} In this dialogue, the user provided information about their departure and arrival city, but have not yet confirmed the date they want to depart. In this case, the assistant should have asked for this information from the user. Instead, the assistant tried to search for buses using a random date.

\subsubsection{Tool-Prediction Error}
\textbf{Definition:} The assistant calls the wrong tool for the task.

\begin{tcolorbox}[breakable, enhanced, colframe=black!75!white, colback=white, sharp corners]
\begin{lstlisting}[language={}]
# Turn 1
    USER: "I need to find a seat on a bus."
    ASSISTANT: "Where are you leaving from? Where are you going?"

# Turn 2
    USER: "I am leaving from Vancouver to go to Seattle."
    ASSISTANT: "When are you leaving?"

# Turn 3
    USER: "I am leaving on the 12th of this month. I need 1 ticket."
    ASSISTANT:
        - API CALL: FindFlights(from_location='Vancouver', leaving_date='2019-03-12', to_location='Seattle', travelers='1')
        - RESULT:
            * {"fare": "239", "leaving_time": "06:40", "departure_airport": "YVR", "arrival_airport": "SeaTac"}
            * {"fare": "331", "leaving_time": "08:10", "departure_airport": "YVR", "arrival_airport": "SeaTac"}
        - RESPONSE: "I found multiple options. The first flight departs at 6:40 am for $29."
\end{lstlisting}
\end{tcolorbox}

\textbf{Explanation:} At turn 3, instead of invoking the FindBus() tool, the assistant called the FindFlights() tool. This tool looks for flights and not bus tickets as requested by the user.

\subsubsection{Required Arguments}
\textbf{Definition:} The assistant makes a mistake in one or more of the required arguments of a function call. Mistakes can range from typos to incorrect values that don't align with the user instructions.

\begin{tcolorbox}[breakable, enhanced, colframe=black!75!white, colback=white, sharp corners]
\begin{lstlisting}[language={}]
# Turn 1
    USER: "I need to find a seat on a bus."
    ASSISTANT: "Where are you leaving from? Where are you going?"

# Turn 2
    USER: "I am leaving from Vancouver to go to Seattle."
    ASSISTANT: "When are you leaving?"

# Turn 3
    USER: "I am leaving on the 12th of this month. I need 1 ticket."
    ASSISTANT:
        - API CALL: FindBus(from_location='Tacoma', leaving_date='2019-03-12', to_location='Vancouver', travelers='1')
        - RESULT:
            * {"fare": "29", "leaving_time": "06:40"}
            * {"fare": "31", "leaving_time": "08:10"}
        - RESPONSE: "I found multiple options. The first bus departs at 6:40 am for $29."
\end{lstlisting}
\end{tcolorbox}

\textbf{Explanation:} At turn 3, the assistant calls the FindBus() tool with the ``from\_location'' argument set to ``Tacoma'' instead of the correct location ``Seattle''. Since ``from\_location'' is a required argument of the function, then this is a required arguments error.

\subsubsection{Optional Arguments}
\textbf{Definition:} The assistant either makes a mistake in one or more optional arguments, omits an optional argument requested by the user, or adds an unnecessary optional argument to the function call.

\begin{tcolorbox}[breakable, enhanced, colframe=black!75!white, colback=white, sharp corners]
\begin{lstlisting}[language={}]
# Turn 1
    USER: "I need to find a seat on a bus."
    ASSISTANT: "Where are you leaving from? Where are you going?"

# Turn 2
    USER: "I am leaving from Vancouver to go to Seattle."
    ASSISTANT: "When are you leaving?"

# Turn 3
    USER: "I am leaving on the 12th of this month. I need 1 ticket."
    ASSISTANT:
        - API CALL: FindBus(from_location='Seattle', leaving_date='2019-03-12', to_location='Vancouver', travelers='1', transfers='0')
        - RESULT:
            * {"fare": "29", "leaving_time": "06:40"}
            * {"fare": "31", "leaving_time": "08:10"}
        - RESPONSE: "I found multiple options. The first bus departs at 6:40 am for $29."
\end{lstlisting}
\end{tcolorbox}

\textbf{Explanation:} At turn 3, the assistant calls the FindBus() tool and sets the ``transfers'' argument to the value $0$. The user did not indicate 
that they care about the number of bus transfers for their trip and might want to consider trips that include transfers. Therefore, the assistant added an additional constraint through this optional argument that was not requested by the user.

\subsubsection{Observation Reasoning}
\textbf{Definition:} The assistant makes a correct tool call but then misinterprets the result of the call and formulates an incorrect or hallucinated response.

\begin{tcolorbox}[breakable, enhanced, colframe=black!75!white, colback=white, sharp corners]
\begin{lstlisting}[language={}]
# Turn 1
    USER: "I need to find a seat on a bus."
    ASSISTANT: "Where are you leaving from? Where are you going?"

...

# Turn 5
    USER: "Book the 6:40 am option."
    ASSISTANT:
        - API CALL: BookBus(from_location='Vancouver', leaving_date='2019-03-12', leaving_time='06:40', to_location='Seattle', travelers='1')
        - RESULT: {"status": "confirmed", "booking_id": "12345"}
        - RESPONSE: "It seems there has been an error in the booking system and your ticket could not be booked. Do you want to try another option?"
\end{lstlisting}
\end{tcolorbox}

\textbf{Explanation:} At turn 5, the assistant makes a correct tool call to book the bus ticket and the tool call returns a positive result indicating that the ticket has been booked. Nevertheless, the assistant incorrectly informs the user that his ticket could not be booked. This response is hallucinated and contradicts the observation from the tool call. 

\subsubsection{Non-Invocation Confirmation}
\textbf{Definition:} The assistant confirms an action was taken without invoking the required tool.

\begin{tcolorbox}[breakable, enhanced, colframe=black!75!white, colback=white, sharp corners]
\begin{lstlisting}[language={}]
# Turn 1
    USER: "I need to find a seat on a bus."
    ASSISTANT: "Where are you leaving from? Where are you going?"

...

# Turn 5
    USER: "Book the 6:40 am option."
    ASSISTANT: "Done. Your ticket has been booked."
\end{lstlisting}
\end{tcolorbox}

\textbf{Explanation:} In this dialogue, after the user asked the assistant to book the specific bus ticket, the assistant responded by confirming to the user that the ticket has been booked. In reality, we see that the assistant did not invoke any tool so no ticket was booked. The response is hallucinated.

\subsubsection{Non-Invocation Hesitation}
\textbf{Definition:} The assistant hesitates and does not call a tool when one is needed.

\begin{tcolorbox}[breakable, enhanced, colframe=black!75!white, colback=white, sharp corners]
\begin{lstlisting}[language={}]
# Turn 1
    USER: "I need to find a seat on a bus."
    ASSISTANT: "Where are you leaving from? Where are you going?"

# Turn 2
    USER: "I am leaving from Vancouver to go to Seattle."
    ASSISTANT: "When are you leaving?"

# Turn 3
    USER: "I am leaving on the 12th of this month. I need 1 ticket."
    ASSISTANT: "Would you like to search for hotel options for when you arrive?"
\end{lstlisting}
\end{tcolorbox}

\textbf{Explanation:} At turn 3, the assistant got all the information needed to call the FindBus() tool to search for bus options for the user. Instead of doing this, the assistant asks the user about hotel preferences. 

\subsubsection{Non-Invocation Hallucination}
\textbf{Definition:} The assistant hallucinates information about some service instead of getting the true information from a tool call.

\begin{tcolorbox}[breakable, enhanced, colframe=black!75!white, colback=white, sharp corners]
\begin{lstlisting}[language={}]
# Turn 1
    USER: "I need to find a seat on a bus."
    ASSISTANT: "Where are you leaving from? Where are you going?"

# Turn 2
    USER: "I am leaving from Vancouver to go to Seattle."
    ASSISTANT: "When are you leaving?"

# Turn 3
    USER: "I am leaving on the 12th of this month. I need 1 ticket."
    ASSISTANT: "I see three bus options. The first one leaves at 6 AM and the fare is $20. The last one leaves at 3 PM and the fare is $25. Let me know what you are intersted in."
\end{lstlisting}
\end{tcolorbox}

\textbf{Explanation:} At turn 3, the assistant should have looked up bus options using the FindBus() tool. Instead, the assistant hallucinated information about three bus options. This information was not the result of any tool call and is thus fake.

\section{Synthetic Data Generation Prompt and Details}
\label{app:generation-prompt}

To train \textit{ToolCritic}, we created a dataset by synthetically inserting errors into error-free dialogues from the SGD dataset. Below, we detail the process, starting with an example of a input dialogue and then explaining how errors were introduced.

\subsection{Demonstration Example}

We start with the following error-free dialogue from the SGD dataset:

\begin{tcolorbox}[breakable, enhanced, colframe=black!75!white, colback=white, sharp corners]
\begin{lstlisting}[language={}, caption={SGD Dialogue Example}]
# Turn 1
    USER: "I need to find a seat on a bus."
    ASSISTANT: "Where are you leaving from? Where are you going?"

# Turn 2
    USER: "I am leaving from Vancouver to go to Seattle."
    ASSISTANT: "When are you leaving?"

# Turn 3
    USER: "I am leaving on the 12th of this month. I need 1 ticket."
    ASSISTANT:
        - API CALL: FindBus(from_location='Vancouver', leaving_date='2019-03-12', to_location='Seattle', travelers='1')
        - Description: Find a bus journey for a given pair of cities
        - Required Arguments:
            * from_location: City where bus is leaving from; is_categorical: False
            * to_location: City where bus is going to; is_categorical: False
            * leaving_date: Date of bus leaving for journey; is_categorical: False
        - Optional Arguments:
            * travelers: Number of travelers for journey; is_categorical: True; Possible Values: ['1', '2', '3', '4', '5']
        - RESULT:
             * {'fare': '29', 'from_location': 'Vancouver', 'from_station': 'Pacific Central Station', 'leaving_date': '2019-03-12', 'leaving_time': '06:40', 'to_location': 'Seattle', 'to_station': 'King Street Station', 'transfers': '0', 'travelers': '1'}
             * {'fare': '31', 'from_location': 'Vancouver', 'from_station': 'Pacific Central Station', 'leaving_date': '2019-03-12', 'leaving_time': '08:10', 'to_location': 'Seattle', 'to_station': 'King Street Station', 'transfers': '0', 'travelers': '1'}
             * {'fare': '26', 'from_location': 'Vancouver', 'from_station': 'Pacific Central Station', 'leaving_date': '2019-03-12', 'leaving_time': '08:30', 'to_location': 'Seattle', 'to_station': 'King Street Station', 'transfers': '0', 'travelers': '1'}
             * {'fare': '30', 'from_location': 'Vancouver', 'from_station': 'Pacific Central Station', 'leaving_date': '2019-03-12', 'leaving_time': '08:50', 'to_location': 'Seattle', 'to_station': 'King Street Station', 'transfers': '0', 'travelers': '1'}
        - RESPONSE: "I found multiple options. First leaves at 6:40 am and is $29. There are 0 transfers."

# Turn 4
    USER: "Anything else available?"
    ASSISTANT: "Next, 8:10 am for $31. Also has 0 transfers."

# Turn 5
    USER: "What is the station I am leaving from? What station am I arriving at?"
    ASSISTANT: "You are leaving from Pacific Central Station and arriving at King Street Station."

# Turn 6
    USER: "Any other options?"
    ASSISTANT: "Next, the bus departs at 8:30 am and the ticket cost is $26. Again, 0 transfers."

# Turn 7
    USER: "That will work, please reserve."
    ASSISTANT: "Please confirm, leaving Vancouver for Seattle on March 12th at 8:30 am. 1 Ticket reserved."

# Turn 8
    USER: "Yes that is correct."
    ASSISTANT:
        - API CALL: BuyBusTicket(from_location='Vancouver', leaving_date='2019-03-12', leaving_time='08:30', to_location='Seattle', travelers='1')
        - Description: Buy tickets for a bus journey
        - Required Arguments:
            * from_location: City where bus is leaving from; is_categorical: False
            * to_location: City where bus is going to; is_categorical: False
            * leaving_date: Date of bus leaving for journey; is_categorical: False
            * leaving_time: Time of bus leaving for journey; is_categorical: False
            * travelers: Number of travelers for journey; is_categorical: True; Possible Values: ['1', '2', '3', '4', '5']
        - Optional Arguments:
        - RESULT: {'fare': '26', 'from_location': 'Vancouver', 'from_station': 'Pacific Central Station', 'leaving_date': '2019-03-12', 'leaving_time': '08:30', 'to_location': 'Seattle', 'to_station': 'King Street Station', 'transfers': '0', 'travelers': '1'}
        - RESPONSE: "Your ticket is confirmed."

# Turn 9
    USER: "Thank you, that will be all."
    ASSISTANT: "Have a great day!"
\end{lstlisting}
\end{tcolorbox}

Notice the format we use to display the conversation. We group each assistant and user message into a single turn. This is to train \textit{ToolCritic} to always expect the last turn to be an assistant turn. Notice also that we include the function specification after each invocation in the middle of the dialogue. This ensures that \textit{ToolCritic} can focus on the tool called and can recall its specification when needed.

\subsection{Hand-Writing Demonstration Examples}

Next, we show how we manually insert an error into this error-free dialogue and provide a detailed description to help guide the LLM in the following few-shot generation process. For this example, we will insert a ``premature invocation'' mistake at Turn 2.

\begin{tcolorbox}[breakable, enhanced, colframe=black!75!white, colback=white, sharp corners]
\begin{lstlisting}[language={}, caption={Demonstration example using the premature-invocation error category.}]
Error Location:
Turn 2

Error Insertion Steps:
1- At Turn 3, locate the API call FindBus(). According to its documentation, this API takes three required arguments ['from_location', 'to_location', 'leaving_date'].
2- The assistant executed this API at Turn 3 after getting all the information (required arguments) from the user to look for bus options.
3- To simulate the premature-invocation error, we will invoke this API in an earlier turn before getting all the needed information (required arguments) from the user.
4- At Turn 2, the user has so far only given information about the 'from_location' and 'to_location' arguments but still hasn't provided information regarding their desired 'leaving_date'.
5- Copy the API call and RESULT from Turn 3 to Turn 2. The assistant's response can also be copied without change.

Explanation (Reasoning):
The user wanted to look for bus options leaving from Vancouver to Seattle. The assistant invoked the FindBus() API. Looking at the documentation for this API, we see that it takes three required arguments ['from_location', 'to_location', 'leaving_date']. At this point in the conversation, the user has not yet provided values for all the required arguments, specifically, the user hasn't yet provided the desired 'leaving_date'. Before searching for bus options using the FindBus() API, the assistant should have obtained more information from the user, specifically about the desired 'leaving_date'. Instead, the assistant invoked the FindBus() API prematurely with some hallucinated argument values, before getting all required info. Therefore, this is a premature-invocation error.

Error Turn:
# Turn 2
    USER: "I am leaving from Vancouver to go to Seattle."
    ASSISTANT:
        - API CALL: FindBus(from_location='Vancouver', leaving_date='2019-03-12', to_location='Seattle', travelers='1')
        - Description: Find a bus journey for a given pair of cities
        - Required Arguments:
            * from_location: City where bus is leaving from; is_categorical: False
            * to_location: City where bus is going to; is_categorical: False
            * leaving_date: Date of bus leaving for journey; is_categorical: False
        - Optional Arguments:
            * travelers: Number of travelers for journey; is_categorical: True; Possible Values: ['1', '2', '3', '4', '5']
        - RESULT:
             * {'fare': '29', 'from_location': 'Vancouver', 'from_station': 'Pacific Central Station', 'leaving_date': '2019-03-12', 'leaving_time': '06:40', 'to_location': 'Seattle', 'to_station': 'King Street Station', 'transfers': '0', 'travelers': '1'}
             * {'fare': '31', 'from_location': 'Vancouver', 'from_station': 'Pacific Central Station', 'leaving_date': '2019-03-12', 'leaving_time': '08:10', 'to_location': 'Seattle', 'to_station': 'King Street Station', 'transfers': '0', 'travelers': '1'}
             * {'fare': '26', 'from_location': 'Vancouver', 'from_station': 'Pacific Central Station', 'leaving_date': '2019-03-12', 'leaving_time
             * {'fare': '26', 'from_location': 'Vancouver', 'from_station': 'Pacific Central Station', 'leaving_date': '2019-03-12', 'leaving_time': '08:30', 'to_location': 'Seattle', 'to_station': 'King Street Station', 'transfers': '0', 'travelers': '1'}
             * {'fare': '30', 'from_location': 'Vancouver', 'from_station': 'Pacific Central Station', 'leaving_date': '2019-03-12', 'leaving_time': '08:50', 'to_location': 'Seattle', 'to_station': 'King Street Station', 'transfers': '0', 'travelers': '1'}
        - RESPONSE: "I found multiple options. First leaves at 6:40 am and is $29. There are 0 transfers."
\end{lstlisting}
\end{tcolorbox}

Note that the above hand-written ``Error Insertion Steps'' will only be used to few-shot the LLM for data generation, but will not be used when fine-tuning ToolCritic. ToolCritic will be fine-tuned to output the error type and the ``reasoning thought'' only.

\subsection{Few-Shot Data Generation Prompt}

After creating multiple demonstration examples like the one above, we used a few-shot prompting approach to instruct an LLM (Claude 3.5 Sonnet) to insert new errors into additional error-free dialogues from the SGD dataset. Below is the system prompt and user prompt used for this task.

\textbf{System Prompt:}
\begin{tcolorbox}[breakable, enhanced, colframe=black!75!white, colback=white, sharp corners]
\begin{lstlisting}[language={}, caption={LLM system prompt for synthetic data generation}]
- You are given a task-oriented dialogue <query> between a user ("USER") and an assistant ("ASSISTANT").
- Your task is to modify the <query> dialogue to simulate a certain type of error made by the assistant, as described in <error-description>, therefore producing a "corrupted" dialogue.
- The dialogue format at each turn follows one of these two options:
  1- If the assistant did not invoke an API during this turn, then the format is as follows:
      # Turn n # The index of the current turn in the conversation containing both USER and ASSISTANT messages.
          USER: # A user message
          ASSISTANT: # The assistant message
  2- If the assistant invoked an API during this turn, then the format is as follows:
      # Turn n # The index of the current turn in the conversation containing both USER and ASSISTANT messages.
          USER: # A user message
          ASSISTANT: # The start of the assistant field
              - API CALL: # The API called by the assistant alongside the argument values. This function returns a RESULT field.
              - Description: # Description of the API
              - Required Arguments: # List of required arguments of the API with their descriptions
              - Optional Arguments: # List of optional arguments of the API with their descriptions
              - RESULT: # The output of the API call. This field will contain a list of option results if the API call was a search/lookup query, or it will contain a dictionary containing the result of an action API such as booking/reservation.
              - RESPONSE: # The final assistant response after observing the output of the API CALL.
- Format your output as a JSON object containing the following four fields: 
    * Error Insertion Steps: Step-by-Step description of how you simulated this error in the dialogue and what changes were made and at which turns.
    * Error Location: The index of the turn where the error was introduced.
    * Explanation: A short paragraph describing how to spot the assistant error, from the perspective of someone reading the dialogue. The description should be step-by-step, in a Chain-of-Thought fashion.
    * Corrupted Dialogue: The updated turn at which the error was inserted.
- You are given a few examples to guide you in <demonstrations>. You are also given a hint in <hint> to help you decide where and how to insert the error.
- Important: 
    - Since all the dialogue turns before the error turn will be kept unchanged, do not return those in the "Corrupted Dialogue" field, only return the turn at which the error was inserted.
    - If there are multiple possible locations to insert an error, choose the one that best matches the error description and that follows the given demonstration examples. If all things are equal, insert the error in the later turn in the dialogue instead of the earlier turns.
    - Only return the JSON Object. Do not include any additional text.
\end{lstlisting}
\end{tcolorbox}

\textbf{User Prompt:}
\begin{tcolorbox}[breakable, enhanced, colframe=black!75!white, colback=white, sharp corners]
\begin{lstlisting}[language={}, caption={LLM user prompt for synthetic data generation}]
Following is the description of the error you are tasked with simulating
<error-description> 
{}
</error-description>

You have the following demonstration examples to guide you
<demonstrations>
{}
</demonstrations>

Now, your task is to modify the following <query> dialogue to introduce the error described above in <error-description>
<query>
{}
</query>

You are given the following hint
<hint>
{}
</hint>

JSON Output:
\end{lstlisting}
\end{tcolorbox}

Notice the \texttt{<hint>} used at the end of the user prompt. The purpose of this hint is to ensure diversity in data generation and avoid model biases. For instance, we observed in our initial experiments that Claude 3.5 Sonnet tends to insert more errors into earlier turns in the dialogues or into tool calls that appeared frequently in the few-shot examples. To mitigate this bias, the hint provides specific guidance, such as the location, which tool argument to focus on, and how to insert the error. The hint is generated by randomly sampling a viable error location from the list of turns and instructing the LLM to insert the error there. For example, if the error category is ``required arguments,'' the hint might look something like: "Focus on the tool call \texttt{ReserveRoundTripFlights()} at Turn 5 and on the argument \texttt{departure date}." This approach ensures a more diverse and representative error dataset.

\section{Supervised Fine-Tuning Prompt and Training Details}
\label{app:training}

\begin{figure*}[t]
\centering
  \includegraphics[width=\linewidth]{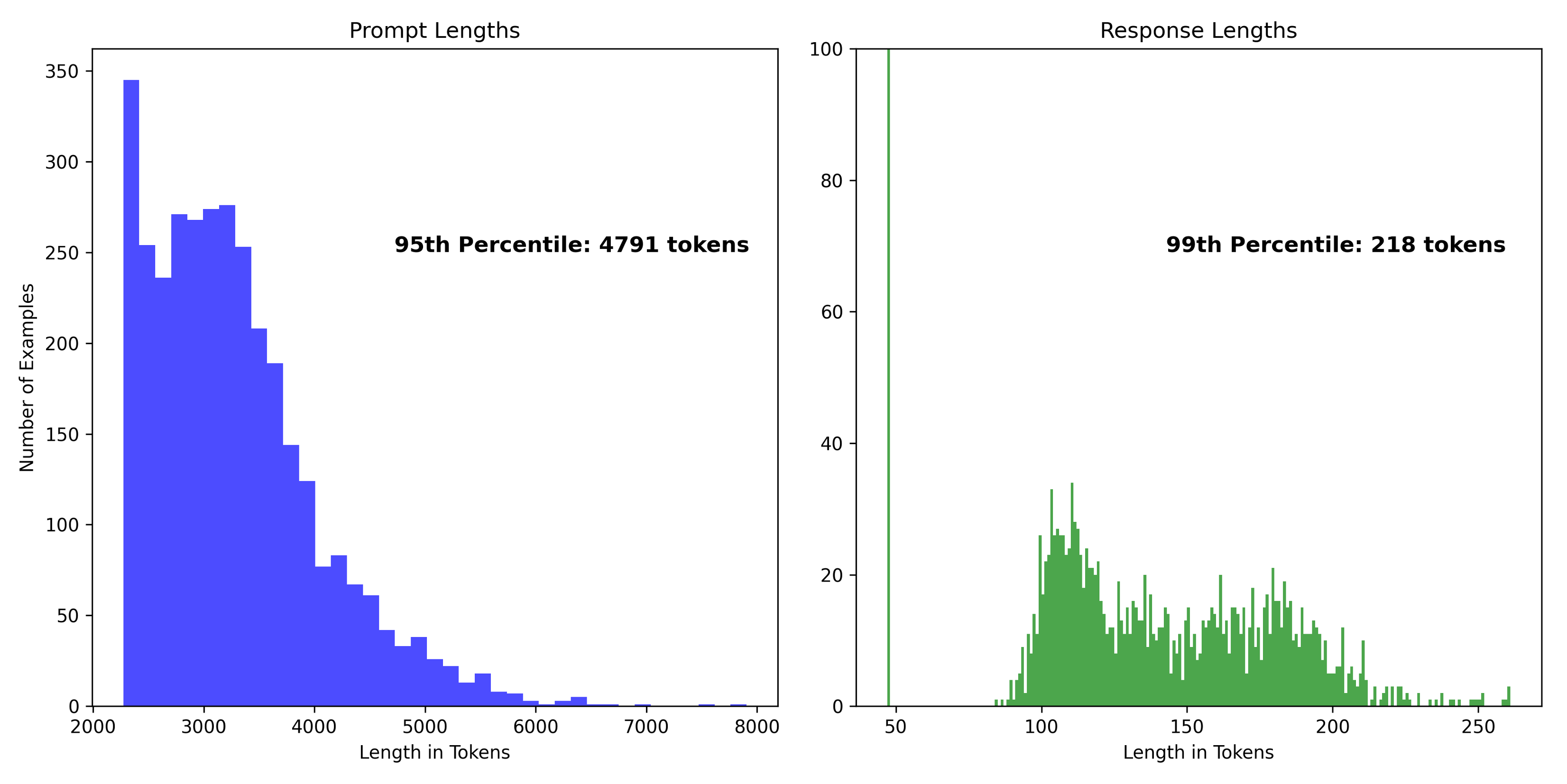}
  \caption {Histogram displaying the length (in tokens) of the input prompt and output response of the fine-tuning training dataset.}
  \label{fig:stats}
\end{figure*}

\subsection{Fine-Tuning Setup}

The fine-tuning process was conducted on the LLaMA 3.1 8B Instruct version, with the following hyperparameters: a learning rate of 1e-5, a batch size of 1, and a temperature of 0.5. The training was performed using bfloat16 precision across 5 epochs, with a cosine learning rate scheduler and a warmup ratio of 0.05. The validation data split was employed exclusively for selecting the learning rate, while all other parameters were kept constant.

We utilized the "DataCollatorForCompletionOnlyLM" function from the \texttt{trl} package~\footnote{See the "DataCollatorForCompletionOnlyLM" from the trl package \url{https://huggingface.co/docs/trl/en/sft_trainer}}, enabling the model to be fine-tuned on completions only. This means that the input prompt and dialogue do not influence the model weights; only the gradients from the response labels are used to update the model. Full fine-tuning was applied, updating all model weights.

To manage token lengths, we limited input prompts to the 95th percentile of the training input prompt lengths, approximately 4971 tokens. The response token length was capped at the 99th percentile of training response lengths, around 218 tokens. Figure~\ref{fig:stats} illustrates the distribution of input and output token lengths. Notably, there is a spike in the response length histogram around 50 tokens, representing the fixed-length response used when the input dialogue is error-free. The standardized response is: "The assistant's final response was appropriate. From the predefined list of error types, the assistant did not commit any errors in the final turn. Therefore, this is a correct turn."

The fine-tuning process adhered to the prompt template style outlined in the official LLaMA documentation, using headers such as \texttt{ASSISTANT\_HEADER} = \texttt{"<|start\_header\_id|> assistant <|end\_header\_id|>"} and \texttt{USER\_HEADER} = \texttt{"<|start\_header\_id|> user <|end\_header\_id|>"}. This format is supported within the Hugging Face library through the \texttt{tokenizer.apply\_chat\_template} function.

\subsection{Fine-Tuning Prompt Template}

The following is the complete fine-tuning prompt template used during training:

\begin{tcolorbox}[breakable, enhanced, colframe=black!75!white, colback=white, sharp corners]
\begin{lstlisting}[language={}, caption={ToolCritic Fine-Tuning Prompt Template}]
Task: Analyze a conversation history between a human user ("USER") and an AI assistant ("ASSISTANT") 
for potential errors in the assistant's final response.

Context:

You will be provided with:
a) A list of API functions available to the assistant in <api-pool>
b) A list of common error types the assistant may make in <error-description>
c) The conversation transcript between USER and ASSISTANT up to a certain turn.

Instructions:

Carefully review the provided API functions, error types, and conversation history.
Analyze the assistant's response in the last turn of the conversation for any of the given error types.
If you detect an error, provide your reasoning for the identified error.
If no errors are detected, simply state that the assistant's response was appropriate.

Please provide your analysis based on the information given in the conversation transcript 
and the provided API and error type descriptions only.

<error_types>
{}
</error_types>

<api_pool>
{}
</api_pool>
\end{lstlisting}
\end{tcolorbox}

Where the \texttt{<error\_types>} field contains the list of all error categories with descriptions, and the \texttt{<api\_pool>} field includes the list of all tool functions available in the dataset. \textit{ToolCritic} requires knowledge of all available tools to accurately detect error types, such as tool-prediction errors and non-invocation errors.

\section{Performance difference on Error Categories in Low-Label Regime}
\label{app:confsuion}

\begin{figure}[h!]
  \includegraphics[width=\linewidth]{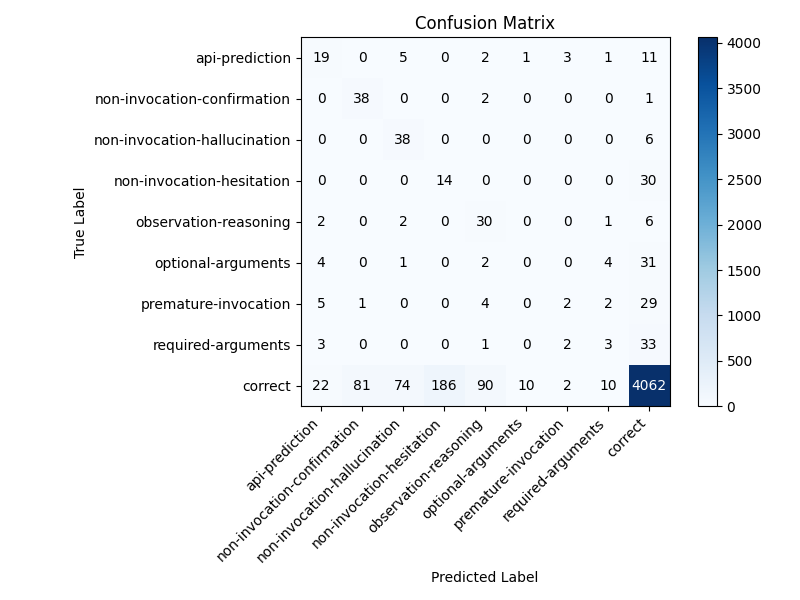} 
  \caption {Confusion matrix on test dataset when training with $10\%$ training data only}
  \label{fig: confusion_low_regime}
\end{figure}

In this section, we analyze the performance of our diagnostic model across different error categories when fine-tuned on a smaller portion of the training set—specifically, $10\%$ of the data, which equates to $22$ examples per error category. Figure~\ref{fig: confusion_low_regime} presents the confusion matrix for the test data split using this fine-tuned model.

From the confusion matrix, it is evident that the model achieves relatively high classification accuracy on the ``non-invocation confirmation'', ``non-invocation hallucination'', and ``observation-reasoning'' error categories. In contrast, for the ``tool-prediction'' (api-prediction) and ``non-invocation hesitation'' categories, the model correctly classifies only about half of the examples. The remaining categories, however, prove to be the most challenging, with the model struggling to detect examples from these error types.

This disparity in performance underscores the difficulty of accurately identifying certain error types in a low-label regime. The results suggest that while the model can generalize well on some categories even with limited data, other categories may require more extensive training data or additional techniques to improve detection accuracy. This could also be an indication that the model is relying on some heuristics to easily detect some error categories. Further data analysis is needed to investigate the reason behind this performance gap, and whether such heuristics can be expected in practice or if this is an artifact of synthetic data generation.

\section{Human Inspection of Revision Quality}\label{app:human-study}

The evaluation in Section~\ref{sec:experiments} has focused on tool-call correctness at the dialogue level. In this subsection, we conduct a human inspection study to examine the more subjective aspects of the dialogue, specifically how feedback impacts the quality of responses at the turn level. This analysis is restricted to the Claude Sonnet model, as it demonstrated the highest baseline performance among the evaluated models, providing a clearer view of how different feedback types influence dialogue quality.

We randomly sampled 100 dialogues from the test set and inspected each turn during evaluation. For each turn, we categorized the impact of the revision into five categories:
\begin{itemize}
    \item \textbf{Useless/No Difference:} The revision made no noticeable improvement. We can think of this as wasting ``compute time''.
    \item \textbf{Made Correct:} The revision resolved an error effectively.
    \item \textbf{Made Better:} The revision improved a response that was already correct, e.g. a smoother or more accurate response.
    \item \textbf{Made Incorrect:} The revision introduced an error into a previously correct response.
    \item \textbf{Missed an Error:} The feedback failed to identify an error.
    \item \textbf{Caught Error but Couldn't Correct:} The feedback identified the error but failed to produce an adequate correction.
\end{itemize}

Table~\ref{tab:manual-inspection} presents the results for the three feedback types: self-reflection, ToolCritic error-only feedback, and ToolCritic full feedback.

\begin{table*}[t]
\centering
\small
\begin{tabular}{lrrr}
\toprule
\textbf{Outcome} & \textbf{Self-Reflection} & \textbf{Full Feedback} & \textbf{Error-Only} \\
\midrule
Useless/No Difference & 40.54\% & 28.89\% & 55.56\% \\
Made Correct          & 13.51\% & 26.67\% & 11.11\% \\
Made Better           & 10.81\% & 26.67\% &  4.44\% \\
Made Incorrect        &  5.41\% &  6.67\% &  6.67\% \\
Missed an Error       & 27.03\% & 11.11\% & 13.33\% \\
Caught Error but Could Not Correct & 2.70\% & 0.00\% & 8.89\% \\
\bottomrule
\end{tabular}
\caption{\label{tab:manual-inspection} 
Results of the human inspection study on Claude 3 Sonnet’s revised responses for 100 sampled dialogues. 
\textit{Full Feedback} produced the largest proportion of beneficial revisions, while \textit{Error-Only} and \textit{Self-Reflection} were often ineffective.}
\end{table*}

\textbf{Analysis:}  
Self-reflection performed poorly, with a high percentage (40.54\%) of revisions deemed useless and a low percentage (13.51\%) that successfully corrected errors. Error-only feedback also showed limited effectiveness, with 55.56\% of revisions being useless and only 11.11\% correcting errors. Full feedback significantly outperformed both methods, with 26.67\% of revisions making the response correct and an additional 26.67\% enhancing the response quality. The percentage of missed errors was lowest for full feedback (11.11\%), and it entirely avoided cases where errors were caught but not corrected.

Overall, the \textit{Full Feedback} strategy yielded substantially more effective revisions (\textit{Made Correct} and \textit{Made Better}) compared to the other methods, confirming that targeted, detailed feedback helps the model produce smoother and more accurate responses. In contrast, \textit{Self-Reflection} and \textit{Error-Only Feedback} frequently produced no discernible improvement or failed to fully address existing errors.

\textbf{Conclusion:}  
In summary, the manual inspection reveals that \textit{ToolCritic (Full Feedback)} is substantially more likely to improve response quality compared to other strategies, highlighting the importance of providing detailed, targeted feedback.

\section{Illustrative Examples of ToolCritic's Generalization Ability}
\label{app:ood}

In this section, we provide illustrative examples to demonstrate the potential generalization capability of \textit{ToolCritic} to new datasets and tools. Specifically, we focus on two examples from the ToolTalk dataset. In both examples, we assume the assistant (Claude 3 Sonnet) has access to a tool called \texttt{AddAlarm} with the following specification:

\begin{tcolorbox}[breakable, enhanced, colframe=black!75!white, colback=white, sharp corners]
\begin{lstlisting}[language={}, caption={Tool Specification from the ToolTalk Dataset}]
{
    "name": "AddAlarm",
    "description": "Add an alarm at a particular time",
    "inputSchema": {
        "json": {
            "type": "object",
            "required": ["time"],
            "properties": {
                "time": {
                            "type": "string",
                            "description": "Alarm Time"
                        },
                "name": {
                            "type": "string",
                            "description": "A name for the alarm"
                        }
            }
        }
    }
}
\end{lstlisting}
\end{tcolorbox}

\subsection{Scenario 1: Non-Invocation Confirmation Error}

In the first scenario, we assume the user asks the assistant to set an alarm at a specific time. To simulate an LLM mistake, we manually write the assistant's response to confirm the alarm has been set, despite no tool being called. The dialogue is as follows:

\begin{tcolorbox}[breakable, enhanced, colframe=black!75!white, colback=white, sharp corners]
\begin{lstlisting}[language={}, caption={Scenario 1 example demonstrating a non-invocation confirmation error}]
# Turn 1
    USER: "Hey I have class tonight at 7. Can you set an alarm for 6:30?"
    ASSISTANT: "Can you please confirm, you want to set an alarm at 6:30 pm?"

# Turn 2
    USER: "Yes"
    ASSISTANT: "Your alarm is now set. Anything else I can help you with?"
\end{lstlisting}
\end{tcolorbox}

We then feed this dialogue to \textit{ToolCritic}, which produces the following output:

\begin{tcolorbox}[breakable, enhanced, colframe=black!75!white, colback=white, sharp corners]
\begin{lstlisting}[language={}, caption={ToolCritic Response with Scenario 1 dialogue as input}]
The user confirmed to the assistant to set an alarm for 6:30 pm. At this point, the assistant  should have invoked the AddAlarm API in order to set the alarm. Instead, the assistant did not  invoke the API but still confirmed to the user that the alarm was set. This is false and hallucinated because the assistant did not call the API. Therefore, this is a non-invocation-confirmation error.
\end{lstlisting}
\end{tcolorbox}

\textit{ToolCritic} correctly identifies this as a "non-invocation-confirmation" error. Interestingly, \textit{ToolCritic} reasons that the \texttt{AddEvent} tool should have been called instead of \texttt{AddAlarm}. This is expected since \textit{ToolCritic} was trained on the tools available in the SGD dataset and has no knowledge of the \texttt{AddAlarm} tool.

Given this feedback and the conversation history, we prompt Claude 3 Sonnet to revise the incorrect response, resulting in the following correct response:

\begin{tcolorbox}[breakable, enhanced, colframe=black!75!white, colback=white, sharp corners]
\begin{lstlisting}[language={}, caption={Claude 3 revised response to Scenario 1 given ToolCritic feedback}]
ASSISTANT:
    - API CALL: AddAlarm(name='Class alarm', time='18:30')
    - RESULT: {'name': 'Class alarm', 'time': '18:30'}
    - RESPONSE: "I've set an alarm named 'Class alarm' for 6:30 PM (18:30) to remind you 
    before your class tonight. Please let me know if you need anything else."
\end{lstlisting}
\end{tcolorbox}

\subsection{Scenario 2: Premature Invocation Error}

In the second scenario, we present a real mistake made by Claude 3 Sonnet according to the following dialogue:

\begin{tcolorbox}[breakable, enhanced, colframe=black!75!white, colback=white, sharp corners]
\begin{lstlisting}[language={}, caption={Scenario 2 example demonstrating a premature invocation error}]
# Turn 1
    USER: "Hey I have an important class tonight evening between 8pm and 10 pm."
    ASSISTANT:
        - API CALL: AddAlarm(name='Important Class', time='20:00')
        - RESULT: {'name': 'Important Class', 'time': '20:00'}
        - RESPONSE: "I've set an alarm named 'Important Class' for 8:00 PM tonight, March 1st, 2019 to remind you of your important class from 8-10 PM. Please let me know if you need anything else!"
\end{lstlisting}
\end{tcolorbox}

In this scenario, the user informed the assistant that they have an important class from 8 PM to 10 PM. An ideal response would have been to ask the user if they wanted to set an alarm and, if so, at what time. Instead, Claude 3 Sonnet sets an alarm at 8 PM without user confirmation. While this response is not entirely incorrect, it could be improved. 

When we feed this dialogue history to \textit{ToolCritic}, we receive the following feedback:

\begin{tcolorbox}[breakable, enhanced, colframe=black!75!white, colback=white, sharp corners]
\begin{lstlisting}[language={}, caption={ToolCritic response with Scenario 2 dialogue as input}]
The user mentioned they have an important class tonight between 8pm and 10pm. The assistant  invoked the AddAlarm API. At this point in the conversation, the user has provided the time  information, but has not yet confirmed that they want to set an alarm. Before adding an alarm using the AddAlarm API, the assistant should have obtained confirmation. Therefore, 
this is a premature invocation error.
\end{lstlisting}
\end{tcolorbox}

\textit{ToolCritic} correctly detects this as a "premature invocation" error, highlighting the assistant's mistake of setting the alarm without user confirmation.

Following this feedback, we prompt Claude again, leading to the following revised response:

\begin{tcolorbox}[breakable, enhanced, colframe=black!75!white, colback=white, sharp corners]
\begin{lstlisting}[language={}, caption={Claude 3 revised response to Scenario 2 given ToolCritic feedback}]
ASSISTANT: "Okay, I understand you have an important class tonight from 8 PM to 10 PM. 
Would you like me to set an alarm to remind you about this class?"
\end{lstlisting}
\end{tcolorbox}

This revised response shows that, with the help of \textit{ToolCritic}'s feedback, Claude 3 Sonnet can produce a more accurate and user-aligned response, ensuring that the assistant's actions align with the user's intent.

\end{document}